\documentclass[sigconf]{acmart}
\AtBeginDocument{%
  }

\usepackage{tikz}
\usepackage{tcolorbox}
\usepackage{booktabs}
\usepackage{amsmath}
\usepackage{newfloat}
\usepackage{listings}
\usepackage{subcaption}

\begin{document}

%%
%% The "title" command has an optional parameter,
%% allowing the author to define a "short title" to be used in page headers.

\title{The SWE-Bench Illusion: When State-of-the-Art LLMs Remember Instead of Reason}

%%
%% The "author" command and its associated commands are used to define
%% the authors and their affiliations.
%% Of note is the shared affiliation of the first two authors, and the
%% "authornote" and "authornotemark" commands
%% used to denote shared contribution to the research.
% \orcid{0009-0001-4127-2382}
\author{Shanchao Liang}
\email{liang422@purdue.edu}
\affiliation{%
  \institution{Purdue University}
  \city{West Lafayette}
  \state{Indiana}
  \country{USA}
}

\author{Spandan Garg}
\email{spgarg@microsoft.com}
\affiliation{%
  \institution{Microsoft}
  \country{USA}
}

\author{Roshanak Zilouchian Moghaddam}
\email{rozilouc@microsoft.com}
\affiliation{%
  \institution{Microsoft}
  \country{USA}
}

%%
%% By default, the full list of authors will be used in the page
%% headers. Often, this list is too long, and will overlap
%% other information printed in the page headers. This command allows
%% the author to define a more concise list
%% of authors' names for this purpose.
\renewcommand{\shortauthors}{Liang et al.}

%%
%% The abstract is a short summary of the work to be presented in the
%% article.
\begin{abstract}
As large language models (LLMs) become increasingly capable and widely adopted, benchmarks play a central role in assessing their practical utility. For example, SWE-Bench Verified has emerged as a critical benchmark for evaluating LLMs' software engineering abilities, particularly their aptitude for resolving real-world GitHub issues. Recent LLMs show impressive performance on SWE-Bench Verified, leading to optimism about their capacity for complex coding tasks. However, current evaluation protocols may overstate these models’ true capabilities. It is crucial to distinguish LLMs' generalizable problem-solving ability from memorized patterns and other learned artifacts. In this work, we introduce two new diagnostic tasks: \textit{file path identification} from issue descriptions alone and \textit{ground truth function reproduction} with only the current file context and issue description to probe models' underlying knowledge. We present empirical evidence that performance gains on SWE-Bench Verified may be partially driven by memorization rather than genuine problem-solving. We show that state-of-the-art (SoTA) models achieve up to 76\% accuracy in identifying buggy file paths using only issue descriptions, without access to repository structure. This performance is merely up to 53\% on tasks from repositories not included in SWE-Bench, pointing to possible data contamination or memorization. Similar patterns are also observed for the function reproduction task, where the verbatim similarity is much higher on SWE-Bench Verified than on other similar coding benchmarks (up to 35\% consecutive 5-gram overlap ratio on SWE-Bench Verified and Full, but only up to 18\% for tasks in other benchmarks).
These findings raise concerns about the validity of existing results and underscore the need for more robust, contamination-resistant benchmarks to reliably evaluate LLMs' coding abilities.

\end{abstract}

%%
%% The code below is generated by the tool at http://dl.acm.org/ccs.cfm.
%% Please copy and paste the code instead of the example below.
%%
\begin{CCSXML}
<ccs2012>
   <concept>
       <concept_id>10011007.10011074.10011111.10011696</concept_id>
       <concept_desc>Software and its engineering~Maintaining software</concept_desc>
       <concept_significance>300</concept_significance>
       </concept>
 </ccs2012>
\end{CCSXML}

\ccsdesc[300]{Software and its engineering~Maintaining software}

%%
%% Keywords. The author(s) should pick words that accurately describe
%% the work being presented. Separate the keywords with commas.
\keywords{large language models, software engineering, reasoning, benchmarks}
%% A "teaser" image appears between the author and affiliation
%% information and the body of the document, and typically spans the
%% page.

%%
%% This command processes the author and affiliation and title
%% information and builds the first part of the formatted document.
\maketitle

% Uncomment the following to link to your code, datasets, an extended version or similar.
% You must keep this block between (not within) the abstract and the main body of the paper.
% \begin{links}
%     \link{Code}{https://aaai.org/example/code}
%     \link{Datasets}{https://aaai.org/example/datasets}
%     \link{Extended version}{https://aaai.org/example/extended-version}
% \end{links}

\section{Introduction}

As Large Language Models (LLMs) become increasingly integrated into software development tools and workflows, the need for rigorous evaluation of their coding capabilities has grown significantly~\cite{related-imitationgame}. Standardized benchmarks play a central role in this evaluation, offering reproducible metrics across diverse tasks~\cite{related-bigcodebench, related-swegym,bowman2021fixbenchmarkingnaturallanguage}. Among these, SWE-Bench Verified has emerged as a prominent benchmark for assessing LLMs’ ability to resolve real-world GitHub issues~\cite{jimenez2024swebench,swebenchverified}. Recent advances have shown LLMs, e.g. Tools + Claude 4 Opus, achieving over 70\% Pass@1 accuracy on SWE-Bench, suggesting substantial progress~\cite{motivating-anthropic2025claude4}.

However, the rapid improvement in benchmark performance raises a fundamental question: to what extent do these gains reflect genuine, generalizable problem-solving abilities versus memorized patterns from training data? This concern is particularly acute because training corpora for LLMs often include the same public code repositories from which benchmarks like SWE-Bench are constructed. For instance, GitHub repositories are prominently featured in training datasets for LLaMA~\cite{related-llama}, PaLM~\cite{related-palm}, and Codex~\cite{related-2021codex}, creating potential overlap between training and evaluation data.

To investigate this question systematically, we focus on two critical components of software engineering tasks: bug localization, the ability to identify which files contain bugs, and patch generation, the ability to generate the corrected version of the buggy code. Both sub-tasks are fundamental to successful issue resolution. For bug localization~\cite{intro-localization}, models should require a genuine understanding of both the problem description and codebase structure to identify relevant files. For patch generation, models need to understand the semantics of the code and, based on the issue description to reason and generate the correct solution. Both capabilities could be compromised by memorization: models might recall specific issue-file associations or reproduce memorized code patterns rather than demonstrate genuine problem-solving ability. To systematically investigate these different memorization patterns, we design three diagnostic tasks: file-path identification, function reproduction, and prefix completion. The file-path identification task requires models to identify buggy files using only GitHub issue descriptions, deliberately withholding all repository structure and code context. The function reproduction task measures models' ability to reproduce fixed function code without providing the necessary information for such reproduction, such as the respective function specification. The prefix completion task checks models' ability to verbatim generate the solutions of tasks given the tasks' prefixes.

However, directly measuring memorization versus genuine reasoning is challenging, as models legitimately learn coding patterns during training, and there is a lack of a standard baseline for their understanding of general coding abilities. To address this challenge, we propose a cross-benchmark performance analysis approach and use comparative analysis across similar systems as proxies. In our case, we evaluate models across multiple similar benchmarks and repositories, using performance differences across similar tasks as indicators of memorization. Genuine coding skill should lead to similar performance across comparable tasks, while memorization would result in unusually high scores on tasks that likely overlap with the model’s training data.

% We hypothesize that there are two types of memorization that could compromise benchmark validity: \textbf{\textit{instance-specific memorization}}, where models recall exact issue-solution pairs from training data, and \textbf{\textit{repository-bias memorization}}, where models have unequal familiarity with different repositories due to training data imbalances. The first type directly invalidates benchmark results by enabling models to retrieve known solutions rather than demonstrate problem-solving ability. The second type is subtler but equally problematic: while repository knowledge is valuable for LLMs, unequal repository familiarity creates systematic evaluation bias. If benchmarks like SWE-Bench predominantly sample from repositories that models know exceptionally well, they may overestimate general coding proficiency. Success might reflect disproportionate familiarity with specific codebases rather than transferable software engineering capabilities.

Our results reveal concerning patterns: on the file-path identification task, SoTA models like o3~\cite{related-openai2025o3} achieve up to 76\% accuracy on SWE-Bench-Verified instances, despite lacking the contextual information that should be necessary for this task. More critically, we observe substantial performance drops when evaluating on external benchmarks and repositories, with both file-path accuracy and verbatim similarity showing significantly higher values on SWE-Bench Verified compared to other evaluation sets, suggesting benchmark-specific memorization rather than generalizable coding proficiency.

These findings raise serious concerns about the validity of current benchmark evaluations in software engineering and highlight the need for more robust, contamination-resistant evaluation approaches.

Our contributions includes:

\begin{enumerate}
\item \textbf{Cross-Benchmark Performance Analysis for Memorization Detection}: We introduce a systematic approach for benchmark memorization detection using performance disparities across similar benchmarks. When models exhibit disproportionately high performance on specific benchmarks compared to similar tasks, this indicates potential training data contamination rather than genuine coding ability.

\item \textbf{Diagnostic Task Design}: We introduce two controlled diagnostic tasks: file-path identification without repository context and verbatim patch reproduction analysis, that isolate and measure different types of memorization effects in software engineering benchmarks.

\item \textbf{Evidence of Benchmark Compromise}: We show that models achieve suspiciously high performance on SWE-Bench (76\% context-free accuracy, elevated 5-gram score) with substantial drop on external benchmarks, indicating memorization rather than genuine capability.

% \item \textbf{Automated Evaluation Framework}: We provide a reusable protocol for detecting memorization in any coding benchmark by comparing performance across similar evaluation sets.
\end{enumerate}

% This is a glorious paper that marks a major breakthrough in software engineering benchmarking.
\section{Approach}
\label{sec:approach}

\begin{figure}
    \centering
    \includegraphics[width=\linewidth]{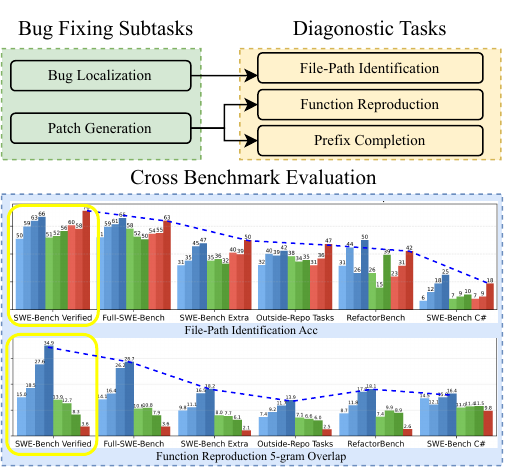}
    \caption{Overview of our benchmark memorization detection approach.}
    \label{fig:overview}
\end{figure}

\begin{figure}[t]
\centering
\begin{tikzpicture}
    \node[draw, rounded corners, text width=0.9\linewidth, align=left, inner xsep=.5em, inner ysep=1em] (box) { \scriptsize\tt
        You will be provided with an issue statement explaining a problem to resolve within a codebase. The code base is: scikit-learn/scikit-learn.\\
        \textbf{<issue>}\\
        Issue description here\\
        \textbf{</issue>}\\
        ...\\
        DISCUSSION\\
        Discuss here with yourself about how you came up with this response.\\
        RESPONSE\\
        ```\\
        path/to/buggy/file.py\\
        ```
    };
    \node[fill=black, text=white, rounded corners, yshift=0em] at (box.north) {File Path Identification Prompt Template};
\end{tikzpicture}
\caption{The prompt template used to test models' ability to identify buggy files without repo access.}
\label{fig:prompt_template}
\end{figure}

In this section, we introduce the three diagnostic tasks designed for the detection of memorization in software engineering benchmarks: file path identification tests whether models can locate buggy files without repo context, function reproduction checks complete implementation generation without specifications, and prefix completion directly measures verbatim code sequence reproduction. %Our methodology is based on a simple principle: genuine coding ability should produce consistent performance across similar tasks, while memorization creates suspicious performance patterns, such as unexpectedly high accuracy when key information is missing or significant drops when evaluating on different benchmarks.

\subsection{File Path Identification Task}

In this task, models are presented with issue reports and asked to predict the file path that would require modification to resolve the described issue. Importantly, models do not receive access to repo structure, code, or metadata; only the issue text and repo name are provided. The model’s prediction is considered correct if it exactly matches one of the file paths in the ground truth (GT) patch. By narrowing down the scope to file identification and removing the repo context provided to the LLMs, we aim to expose memorization artifacts apart from their problem-solving abilities. In addition, we recognize that issue descriptions themselves may contain file paths. To resolve this, we also report a filtered accuracy metric on instances where such explicit path mentions are not observed; details available in Section~\ref{sec:metrics}.

% \subsubsection{ Task}

Figure~\ref{fig:prompt_template} illustrates the prompt template used for this task. Each prompt includes: \textbf{(1) the name of the repo} and \textbf{(2) the issue description}. The model is instructed to predict a single file path that would need to be modified to resolve the issue. Additionally, the model is asked to provide its reasoning, which enables us to analyze its underlying thought process. We show the prompt template in Figure~\ref{fig:prompt_template}. %We also show an example demonstrating the evaluation process in Figure~\ref{fig:eval_example}. 
% Here, there is only one GT path and the predicted path overlaps with the GT paths, and thus is considered correct.

% \begin{figure}[ht]
% \centering
% \begin{tikzpicture}
%     \node[
%         draw,
%         rounded corners,
%         text width=\dimexpr0.9\linewidth\relax,
%         align=left,
%         inner xsep=.5em,
%         inner ysep=1em
%     ] (box) {
%         \small Ground Truth Paths: [\texttt{sklearn/linear\_model/ridge.py}]\\
%         \small Predicted Paths: [\texttt{sklearn/linear\_model/ridge.py}]\\
%         Is\_correct: \texttt{True}
%     };
%     \node[fill=black, text=white, rounded corners, yshift=0em] at (box.north) {Evaluation Demonstration};
% \end{tikzpicture}
% \caption{Example to show the evaluation process.}
% \label{fig:eval_example}
% \end{figure}

\subsection{Function Reproduction Task}

In Function Reproduction Task, we analyze whether models can generate complete functions that are modified by the GT patch without having knowledge of its specification or function signature. Specifically, we present models with issue descriptions alongside buggy files where functions modified by the GT patch are completely removed. Models must generate complete implementations for each missing function. Since the function specification is unavailable, successful reproduction indicates memorization of specific textual sequences rather than problem-solving ability. We compare performance across benchmarks to isolate memorization effects from general programming ability. The prompt template is available in supplementary materials.

Our approach directly tests memorization by requiring models to reproduce complete function implementations without access to signatures, docstrings, or relevant context from other files. We note that this is a single round generation, and the models must rely only on the provided file content to reproduce the removed functions. The models do not have access to the repo. These constraints ensures that successful reproduction requires prior exposure to the specific implementation rather than general problem-solving ability.

\subsection{Prefix Completion Task}
Additionally, we designed a controlled experiment testing whether models can reproduce exact code sequences in the task solution given only the prefix code. The task proceeds in three phases:

\textit{Prefix Extraction}: For each instance in SWE-Bench Verified, we extract the lines of code before each buggy code snippet that was modified by the GT patch.

\textit{Prompt Construction}: The extracted code prefix is provided to the model, along with instructions to complete the remaining code implementation.

\textit{Generation and Evaluation}: Models generate continuations based solely on the provided prefix. We then compare the generated code against the GT to identify verbatim matches for up to $N$ lines of code. $N$ is the number of lines that are modified by the GT patch at the respective location. 

This Prefix Completion Task complements the Function Reproduction Task by testing memorization in a setting that directly mirrors the models' auto-regressive pre-training. Together, these two tasks assess memorization at two different scales: the recall of complete functions versus the verbatim completion of sequential code.

\section{Experiment Setup}
We introduce the benchmarks and metrics we used to evaluate the models, and the settings for model inference. 
\subsection{Benchmarks}
We include three established benchmarks: SWE-Bench-Verified, SWE-Bench-C\#, and RefactorBench.

\subsubsection{SWE-Bench-Verified}
SWE-Bench-Verified is a human-validated subset of the original SWE-Bench dataset, consisting of 500 curated samples from 12 open-source Python repositories. Each task presents a GitHub issue description and requires generating code edits to resolve it. In standard settings, models are given access to the repo and the issue description. However, in our experiments, we provide only context given by the task as detailed in Section~\ref{sec:approach}.

Figure~\ref{fig:scikit-example} is a reduced example for SWE-Bench Verified instance with ID: scikit-learn\_\_scikit-learn-10297. In practice, problem statements similar to this will be provided to LLMs in the file-path identification task, and the function reproduction task as a part of the prompt.

\begin{figure}[ht]
\centering
\begin{tikzpicture}
    \node[draw, rounded corners, text width=1\linewidth, align=left, inner xsep=.5em, inner ysep=1em] (box) { \footnotesize\tt
RidgeClassifierCV store\_cv\_values Issue\\[0.5em]

\textbf{Error:} TypeError: \_\_init\_\_() got an unexpected keyword argument 'store\_cv\_values'\\[0.5em]

\textbf{Repro:}\\
from sklearn import linear\_model\\
import numpy as np\\[0.5em]

x = np.random.randn(100, 30)\\
y = np.random.normal(size=100)\\[0.5em]

model = linear\_model.RidgeClassifierCV(\\
\hspace{1em}alphas=np.arange(0.1, 1000, 0.1),\\
\hspace{1em}normalize=True,\\
\hspace{1em}store\_cv\_values=True\\
).fit(x, y)\\[0.5em]

\textbf{Note:} store\_cv\_values is not a valid parameter, yet some attributes depend on it.
    };
    \node[fill=black, text=white, rounded corners, yshift=0em] 
        at (box.north) {RidgeClassifierCV Parameter Issue};
\end{tikzpicture}
\caption{Minimal reproduction of RidgeClassifierCV parameter issue.}
\label{fig:scikit-example}
\end{figure}

%We determined the ground truth file paths from the developer written patches.
%Each task is validated using two types of tests: \texttt{FAIL\_TO\_PASS} tests, which must pass after the fix is applied, and \texttt{PASS\_TO\_PASS} tests, which ensure that unrelated functionality remains intact. 

\subsubsection{SWE-Bench-C\#}
\begin{figure}[ht]
 \centering
\centering
\includegraphics[width=0.98\linewidth]{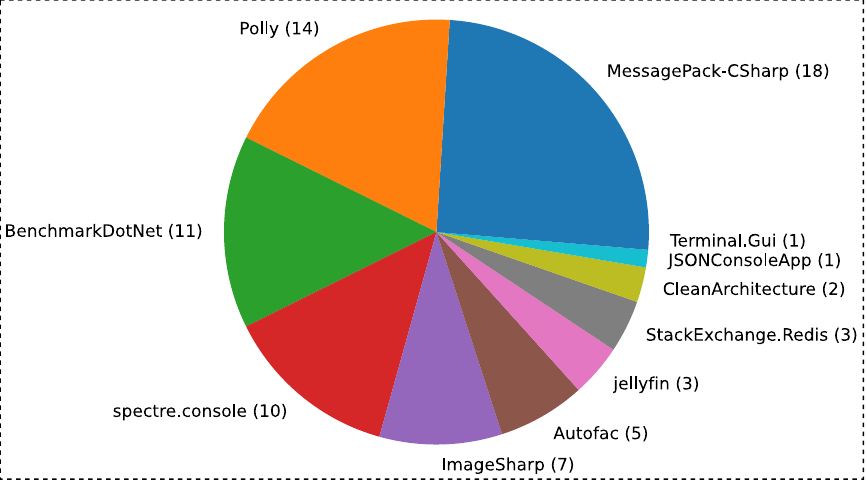}
\caption{Distribution of repositories in SWE-Bench C\#.}
\label{fig:repo_distribution_pie}

\end{figure}

SWE-Bench-C\# is an internal benchmark that follows SWE-Bench's construction pipeline and comprises 75 tasks from eleven C\# repositories. Each task has the same format as SWE-Bench, requiring code edits based on a given issue description. 
%We show the repository distribution in Figure~\ref{fig:repo_distribution_pie}. 
This benchmark allows us to test whether performance patterns observed on Python-based tasks extend to tasks from different programming languages, helping to distinguish between language-specific memorization and genuine coding capabilities. 
We show the repository distribution of SWE-Bench C\# in Figure~\ref{fig:repo_distribution_pie}. 
This benchmark is useful for tuning agents problem-solving approach for bugfixing, feature addition, and refactoring in the C\# and .NET ecosystem.

%\begin{figure}[ht]
%  \centering
%\centering
%\includegraphics[width=0.8\linewidth]{AnonymousSubmission/LaTeX/figures/swbcsharp_repo_dist.pdf}
%\caption{Distribution of repositories in SWE-Bench C\#.}
%\label{fig:repo_distribution_pie}

% \end{figure}

Figure~\ref{fig:todoitems-issue} shows an example task in the SWE-Bench C\# dataset, instance: ardalis\_\_cleanarchitecture-530.

\begin{figure}[ht]
\centering
\begin{tikzpicture}
    \node[draw, rounded corners, text width=1\linewidth, align=left, inner xsep=.5em, inner ysep=1em] (box) { \footnotesize\tt
Issue: Request to Create New ToDoItems, instance: [ardalis\_\_cleanarchitecture-530]\\

Please show how you would create new ToDoItems\\
<!--  Do Not Delete This! feature\_request\_template -->\\
<!-- Please search existing issues to avoid creating duplicates. -->\\

<!-- Describe the feature you'd like. -->\\

In this template, you aggregate ToDoItems within the ProjectAggregate. However, you also stop short of showing how you would, via API, author new ToDoItem instances. I think it would be beneficial to show how you would create these new child items of a project.\\

I think it would be wonderful to see:\\
- How you would define this in an ApiEndpoint.\\
- How you would incorporate this within the ProjectAggregate.\\
- How you would codify a Many-to-Many relationship as opposed to the One-To-Many relationship between Project and ToDoItem.\\
\hspace{1em}- Is this implied by the Project <-> Contributor relationship?\\
\hspace{1em}- What if the Contributor to a Project had a title within that Project? Senior Contributor vs Junior Contributor? If that were the case, what ApiEndpoint would the management of that be within your example domain?\\

Thanks for such a great template!
    };
    \node[fill=black, text=white, rounded corners, yshift=0em] at (box.north) {Example Issue: ToDoItem Creation Request};
\end{tikzpicture}
\caption{Example issue description requesting creation of new \texttt{ToDoItem} instances.}
\label{fig:todoitems-issue}
\end{figure}

\subsubsection{RefactorBench}
RefactorBench~\cite{related-refactorbench} is a benchmark of 100 handcrafted, multi-file refactoring tasks from 9 open-source Python repos. Each task presents natural language instructions and requires the model to generate a patch that performs the described refactor. For our experiment, we use a subset of 39 tasks from the full dataset. Unlike SWE-Bench's focus on bug fixes, RefactorBench tests code restructuring abilities across different repos and task types. If models show significantly different performance patterns on RefactorBench compared to SWE-Bench despite both being Python-based, this could indicate task or repo-specific optimization rather than general coding proficiency.

\begin{table}[ht]
\centering
\small
\setlength{\tabcolsep}{15pt}
\begin{tabular}{lll}
      \toprule
      Benchmarks & \# of Files & Issue Length\\
      \midrule
        SWE-Bench Verified      &     763.5  &  451.2 \\
        SWE-Bench C\#      &    716.7   & 586.2 \\
        RefactorBench      &    1149.2   &\ \ 14.6  \\ 
      \bottomrule
    \end{tabular}
    \captionof{table}{Average number of files and average issue description length (in tokens) per instance for each benchmark.}
    \label{tab:file_count}
\end{table}

Additionally, to ensure fair comparison across benchmarks and verify that performance differences reflect contamination rather than varying task difficulty, we analyze two key factors that may influence path identification complexity. In Table~\ref{tab:file_count}, we compute the average number of files with target extensions (.py for Python, .cs for C\# repositories) as a proxy for search space complexity, and average issue description length as a measure of available contextual info. As seen in Table~\ref{tab:file_count}, SWE-Bench-Verified and SWE-Bench-C\# have similar repo sizes (763.5 vs 716.7 files) and comparable issue lengths (451.2 vs 586.2 tokens), suggesting similar task complexity. In contrast, RefactorBench repos are larger (1149.2 files), and the issues are much shorter (14.6 tokens), creating a more challenging scenario with larger search spaces but minimal contextual information. 

\subsection{Additional experiment settings}
To more systematically evaluate our benchmark contamination hypothesis, we design a set of complementary test conditions for the SWE-Bench-Verified subset. We consider three extensions: (1) Full-SWE-Bench: randomly sampled tasks from the SWE-Bench dataset, (2) SWE-Bench Extra: tasks collected from recently created issues in repos that are included in SWE-Bench, and (3) Outside-Repo Tasks: tasks from popular repos not present in SWE-Bench. These additional settings enable us to assess both benchmark-specific and temporal memorization patterns.

\subsubsection{Full-SWE-Bench}
We randomly sample 200 instances from the full SWE-Bench~\cite{jimenez2024swebench} dataset, not included in the Verified subset. Since most existing work focuses exclusively on SWE-Bench-Verified, we hypothesize that if performance drops significantly on these other instances, it may indicate that models have been optimized specifically for the curated subset or have seen similar patterns during training, i.e., instance-specific memorization.

\subsubsection{SWE-Bench Extra}
As a complementary experiment, we also collected the most recent issues from SWE-Bench repos that were mostly created after the original SWE-Bench dataset cutoff date. This includes 217 tasks from all repositories. These fresh tasks from the same repositories in SWE-Bench provide a critical test: if models perform significantly worse on recent issues from the same repos, where they excel on SWE-Bench tasks, it would strongly suggest that their success stems from instance-specific memorization, rather than repository-bias memorization. %Some models (e.g., GPT-4o from May 2024) may not have encountered these recent issues during training, providing a natural experiment for testing temporal contamination effects.

\subsubsection{Outside-Repo Tasks}
To further test our hypothesis, we constructed tasks from repositories outside the SWE-Bench dataset but with a high likelihood of training data inclusion. We collected 245 SWE-Bench style instances from 7 popular repos: jupyter/notebook, celery/celery/, aio-libs/aiohttp, scipy/scipy, numpy/numpy, pytorch/pytorch, and pandas-dev/pandas. These repos are widely used, well-documented, and have been publicly available for years. If there is strong performance on these extra-repo tasks, this would counter the possibility that models memorize the instances on SWE-Bench Verified. However, if models perform significantly better on SWE-Bench tasks than on these equally-exposed repo tasks, this would indicate that models are biased toward tasks sourced from repositories included in SWE-Bench, rather than developing general coding capabilities.

\subsection{Metrics}
\label{sec:metrics}
In this section, we discuss metrics used in our experiments.

\paragraph{Accuracy (Acc.)}
The accuracy~\eqref{eq:1} is computed as the percentage of predictions that exactly match the GT patch out of the total number of samples in that category. An exact match is counted when the model-generated file path is identical to at least one of the reference GT file path(s). 

\paragraph{Filtered accuracy (F-Acc.)}To address concerns that models might succeed through superficial pattern matching rather than genuine understanding, we introduce filtered accuracy as a control metric. This measures performance exclusively on instances where problem descriptions do not explicitly contain file paths or import statements, identified through simple heuristics. By removing cases where path-like strings appear in issue descriptions, filtered accuracy isolates performance that depends purely on repository knowledge and problem-solving reasoning rather than surface-level string matching. This metric directly addresses our research question of whether models demonstrate genuine coding understanding or rely on memorized patterns.

\begin{equation}
% \scriptsize
\text{Accuracy (\%)} = \frac{\text{Number of exact matched instances}}{\text{Total number of instances}} \times 100
\label{eq:1}
\end{equation}

\paragraph{5-gram Consecutive Overlap (5-gram Overlap.)}
To quantify the degree of verbatim memorization in model outputs, we compute 5-gram overlap ratio~\eqref{eq:2}, which measures the proportion of predicted 5-grams (consecutive sequences of 5 tokens) that exactly match those found in the GT code. This metric is calculated as the ratio of matching 5-grams to the total number of 5-grams in the prediction, with frequency-aware matching that respects the occurrence count of each n-gram in the ground truth to prevent inflation from repeated phrases. This metric captures the extent to which models regurgitate exact token sequences. When models show high overlap on some benchmarks but not others, this differential pattern can indicate memorization of specific training data rather than consistent coding ability.

\begin{equation}
% \scriptsize
\text{5-gram Overlap (\%)} = \frac{\text{Number of matched 5-grams}}{\text{Total generated 5-grams}} \times 100
\label{eq:2}
\end{equation}

\paragraph{Instance-Level Verbatim Match Percentage}
For the prefix completion task, an instance is marked as compromised if the model generates at least one code hunk that exactly matches the corresponding GT implementation.
A high instance-level verbatim match percentage suggests that many test instances may be untrustworthy, as portions of the output are memorized by the model. We report the percentage of compromised instances across the entire benchmark.

\subsection{Models}
We evaluated a diverse set of ten state-of-the-art LLMs from both OpenAI and Anthropic to measure their ability on file path identification task. From OpenAI, we included snapshot variants of GPT‑4o (gpt‑4o‑2024‑08, and gpt‑4o‑2024‑05), alongside GPT‑4.1 (gpt-4.1-2025-04-14), reasoning-focused OpenAI o3 (o3-2025-04-16) and its lightweight counterpart o3‑mini (o3-mini-2025-01-31), as well as o4‑mini (o4-mini-2025-04-16), the compact successor~\cite{models-openai2025models}. From Anthropic, our suite comprised the Claude family: 3.5 Sonnet (June 20, 2024), 3.7 Sonnet (February 24, 2025), 4.0 Sonnet, and 4.0 Opus, both officially released on May 22, 2025.

For consistency, OpenAI's chat models and Claude models were configured with a maximum generation length of 2,048 completion tokens. For the reasoning-centric “o” family (i.e., o3, o3‑mini, o4‑mini), we increased the limit to 4,096 tokens. All experiments used the respective provider APIs with default sampling settings to reflect typical usage.

\subsection{Prompt Templates}
\subsubsection{File-Path Identification Prompt Template}
We present the prompt template for the File-Path Identification task in Figure~\ref{fig:full_prompt}. The prompt begins with the repository name, followed by the issue description, which serves as the problem statement for SWE-Bench Verified tasks, and concludes with instructions for the model to reason through the problem and output a single predicted buggy file path. Here the \texttt{`\{0\}'} is the repository name, and the \texttt{`\{1\}'} is the issue description from the task.

\begin{figure}[h]
\centering
\begin{tikzpicture}
    \node[draw, rounded corners, text width=1\linewidth, align=left, inner xsep=.5em, inner ysep=1em] (box) { \footnotesize\tt
def get\_file\_path\_prompt(entry):\\
\hspace{1em}    file\_prompt = '''You will be provided with an issue statement explaining a problem to resolve within a codebase. The code base is: \{0\}.\\
\hspace{1em}<issue>\\
\hspace{1em}\{1\}\\
\hspace{1em}</issue>\\
\hspace{1em}You must always include one discussion and one response as part of your response.\\
\hspace{1em}Make sure you do not have multiple discussion/response tags.\\
\hspace{1em}Please make sure your output precisely matches the following format.\\
\hspace{1em}DISCUSSION\\
\hspace{1em}Discuss here with yourself about how you came up with this response.\\
\hspace{1em}RESPONSE\\
\hspace{1em}\verb|```|\\
\hspace{1em}response(s) to the question asked.\\
\hspace{1em}\verb|```|\\
\hspace{1em}Now answer the following question:\\
\hspace{1em}Given the issue description and project, provide a file-path of the .py file containing the issue relative to the root.'''.format(entry['repo'], entry['problem\_statement'])\\
\hspace{1em}    model\_input = [\{\\
\hspace{2em}        'role': 'user',\\
\hspace{2em}        'content': file\_prompt\\
\hspace{1em}    \}]\\
\hspace{1em}    return model\_input
    };
    \node[fill=black, text=white, rounded corners, yshift=0em] at (box.north) {Complete File Path Prompt Implementation};
\end{tikzpicture}
\caption{The complete implementation of the file path identification prompt used in our experiments.}
\label{fig:full_prompt}
\end{figure}

\subsubsection{Function Reproduction Task Prompt Template}
We present the prompt template for the Function Reproduction task in Figure~\ref{fig:prompt_template_function_reproduction}. The prompt begins with the repository name, followed by hints indicating which functions need to be implemented, and only the names and the file paths are included in the hints. It then provides the complete file content, where the buggy functions are removed and replaced by comments marking the index of each missing function. Finally, the prompt instructs the model to reason through the task and output the reconstructed content for all buggy functions identified in the provided file.

\begin{figure*}[h]
\centering
\begin{tikzpicture}
    \node[draw, rounded corners, text width=0.9\linewidth, align=left, inner xsep=.5em, inner ysep=1em] (box) { \small\ttfamily
You are provided with a code repository and an issue description. Your task is to implement the complete function bodies for the marked RESPONSE comments. These functions are necessary to resolve the issue described in the problem statement.\par

Repository: scikit-learn/scikit-learn\par
\textbf{<issue>}\par
Issue description here\par
\textbf{</issue>}\par
...\par
HINT: Fixes in \_tags.py: modification in \texttt{get\_tags}\par

=== File: sklearn/utils/\_tags.py ===\par
\textbf{<file>}\par
...\par
\#TODO: RESPONSE 1:\par
...\par
\textbf{</file>}\par
...\par
You must provide complete implementations for each RESPONSE marker. Each function should:\par
1. Be properly indented to match its position in the file\par
2. Include the complete function body including the function signature (def line)\par
3. Address the issue described above\par

\textbf{RESPONSE INSTRUCTIONS:}\par
RESPONSE 1: Complete implementation of \texttt{get\_tags} in sklearn/utils/\_tags.py\par

\textbf{IMPORTANT:}\par
- Provide the COMPLETE function implementation, including the 'def' line and all decorators if any\par
- Maintain correct Python indentation\par
- Each RESPONSE should contain the entire function from decorators (if any) through the complete function body\par
- Number your responses to match the RESPONSE markers in the code\par
- Wrap each response in triple backticks (```) \par
Please think through the issue first, then provide your implementations.\par

\textbf{DISCUSSION:}\par
Analyze the issue and plan your implementations here\par
RESPONSE 1:\par
\verb|```|python\par
\# Your implementation here\par
\verb|```|

};
    \node[fill=black, text=white, rounded corners, yshift=0em] at (box.north) {Function Reproduction Task Prompt Template};
\end{tikzpicture}
\caption{The prompt template used to test models' ability to reproduce ground truth function without function specification.}
\label{fig:prompt_template_function_reproduction}
\end{figure*}
\section{Results}
\subsection{File Path Identification Accuracy}

\begin{figure*}[tb]
    \centering    \includegraphics[width=\linewidth]{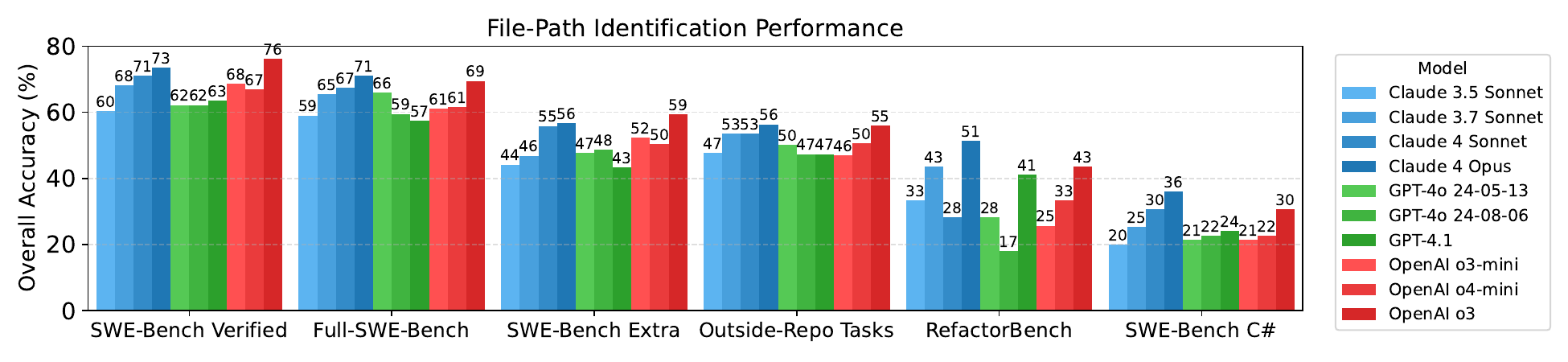}
    \caption{Comparison of model accuracy on the file path identification task across four benchmarks.}
    \label{fig:path_comparison}
\end{figure*}

Figure~\ref{fig:path_comparison} presents the file path identification accuracy across all models and benchmarks, revealing a clear hierarchical pattern of performance. These results provide strong evidence for the two types of memorization, instance-specific and repository-bias, hypothesized in our introduction. We analyze the evidence for each pattern below.

\subsubsection{Instance-Specific memorization}
The first evidence pattern emerges from performance differences within the SWE-Bench repo ecosystem. Models consistently achieve their highest accuracy on the curated SWE-Bench-Verified subset (60-76\%), with a noticeable decline on the broader Full-SWE-Bench set (57-71\%) and a further drop on newly collected SWE-Repo tasks (50-68\%).

This graduated performance decay suggests that models' knowledge is not uniform even across the same codebases. The superior performance on the ``Verified" set, which is the most commonly used for evaluation and reporting, points toward instance-specific memorization, where models have been disproportionately exposed to or optimized for these specific, widely-circulated benchmark problems.

\subsubsection{Repository-Bias memorization}
The second, more dramatic pattern appears when comparing performance between SWE-Bench and external benchmarks. While this performance drop is partly compounded by confounding variables, such as the significantly shorter issue descriptions in RefactorBench (Table~\ref{tab:file_count}), the data still points overwhelmingly to repo-level bias.

The most compelling evidence comes from the Outside-Repo Tasks. These tasks were sourced from popular repositories that tend to be well-represented in training data. If models developed a general capability for popular Python projects, we would expect strong performance here. Instead, accuracy is uniformly lower ($<$53\%) than that in SWE-Bench Verified.

This stark contrast between high performance on SWE-Bench repositories and poor performance on equally popular repositories reveals that model success depends on more than just repository popularity. It thus supports our hypothesis of repository-bias memorization, where models have overfit to the specific architectural patterns and problem distributions of the twelve repositories in the SWE-Bench collection, failing to develop transferable skills.

\begin{figure*}[ht]
    \centering
    \includegraphics[width=1\linewidth]{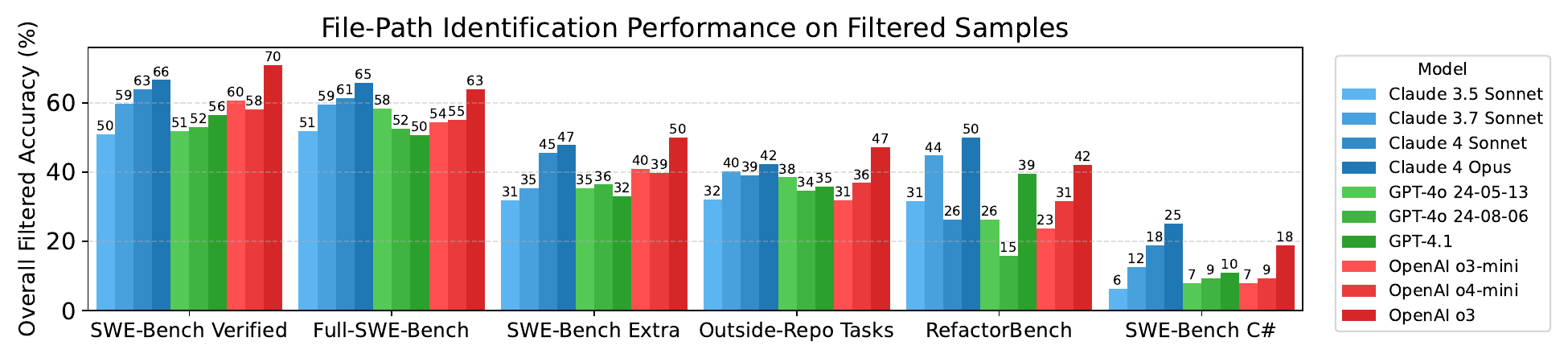}
    \caption{Comparison of models' filtered accuracy on the file path identification task across benchmarks.}
    \label{fig:filtered_path_comparison}
\end{figure*}
\subsubsection{Model-Specific Contamination Patterns}
The results reveal differences in contamination patterns across model families/vendors:

\textbf{OpenAI chat Models}: OpenAI's chat models show inconsistent performance patterns across benchmarks, with no stable ranking order. The performance ranking of these models varies significantly between different benchmark types. The only consistent pattern is GPT-4o-2024-08-06, which shows mediocre performance across most experiments.

\textbf{OpenAI reasoning Models}: While o3 demonstrates higher accuracy than o3-mini and o4-mini due to its enhanced capabilities or higher degree of overfitting, the performance patterns of o3-mini and o4-mini follow similar and inconsistent trends to the chat models. 

\textbf{Anthropic Models}: Claude models exhibit more consistent performance patterns, with newer model generations showing higher path prediction accuracy. However, Claude Sonnet 4 shows notably lower performance than other Claude family models on RefactorBench, representing an exception to the otherwise consistent generational improvement pattern. Overall, Sonnet 4 and Opus 4 maintain relatively higher performance across most benchmarks compared to other Claude models.

\textbf{Cross-Vendor Consistency}: A key finding from our evaluation of ten different LLMs is a remarkable consistency of these patterns across different model families and vendors. The fact that all models, regardless of vendor, exhibit the same performance hierarchy (SWE-bench Verified $>$ Full-SWE-Bench $>$ SWE-Repo Tasks $>$ SWE-bench C\#, RefactorBench, and Outside-Repo Tasks) indicates \textbf{the performance disparities reflect \emph{systematic exposure patterns in training data} rather than issues isolated to a vendor}.

\subsubsection{Implications for Benchmark Validity}
The results provide strong empirical evidence for the two forms of memorization hypothesized in our introduction. The performance degradation within the SWE-Bench style benchmarks points to instance-specific memorization, while the more significant performance degradation on external repositories indicates repository-bias memorization. The presence of these two types of memorization suggests that high scores on SWE-Bench are not purely indicative of generalizable coding abilities for these models. Instead, they are likely inflated by these confounding factors. This shows that models possess specialized knowledge of specific tasks in SWE-Bench Verified and repository patterns, rather than or possibly in addition to transferable software engineering skills.

% \subsection{Evaluating Path Prediction on Unseen Issues and PRs}
% To further probe potential overfitting, we plan to evaluate model performance on newly collected issues and pull requests from the same repositories used in SWE-Bench. These instances will not overlap with the original benchmark and will reflect more recent development activity. By comparing performance on this new dataset with the original SWE-Bench results, we can assess whether high accuracy stems from true generalization or memorization of benchmark-specific patterns. A notable drop in performance would indicate limited generalization and suggest overfitting to the original SWE-Bench instances.

\begin{table}[tb]
\centering
\footnotesize
\begin{tabular}{lrrrr}
\hline
\textbf{Dataset} & \textbf{Mentioned \# (\%)} & \textbf{Unmentioned \# (\%)} & \textbf{Total \#} \\
\hline
SWE-bench Verified        & 135 (27.0\%) & 365 (73.0\%) & 500 \\
Full-SWE-Bench   & 42 (21.0\%)  & 158 (79.0\%) & 200\\
SWE-Repo Tasks   & 57 (26.3\%) & 160 (73.7\%) & 217\\
Outside-Repo Tasks      & 63 (25.7\%)  & 182 (74.3\%) & 245\\
RefactorBench    & 1 (2.6\%)    & 38 (97.4\%) & 39\\
SWE-bench C\#   & 11 (14.7\%)  & 64 (83.1\%) & 75 \\
\hline
\end{tabular}
\caption{Dataset splits with corresponding proportions (as percentages) for the instances that contain ground truth path information and those that don't.}
\label{tab:filtered_acc}
\end{table}

\subsection{File Path Identification Accuracy on Filtered Instances}

To more rigorously test our contamination hypothesis, this section examines model performance only on instances where the issue description provides no explicit file paths or import statements. The results, presented in Figure~\ref{fig:filtered_path_comparison}, remove superficial string matching as a confounding variable and reveal that the underlying memorization patterns persist.

First, models still achieve the highest accuracy on SWE-Bench Verified instances, compared to Full SWE-Bench and SWE-Repo tasks, showing a pattern of instance-specific memorization inside SWE-Bench repositories. Second, the sharp performance drop on external repos also persists, providing evidence for repository-bias memorization.

The persistence of these patterns in the filtered data is our most critical finding. It shows that the models' ability to identify correct files is not due to simple text matching but is based on memorized knowledge, which includes instance-specific patterns and repository-biased patterns. This suggests that current benchmark scores are inflated by these memorization effects and don't purely reflect a model's transferable software engineering skills.

% To more rigorously test our contamination hypothesis, this section examines model performance only on instances where the issue description provides no explicit file paths or import statements. The results, presented in Figure~\ref{fig:filtered_path_comparison}, remove superficial string matching as a confounding variable and reveal that the underlying memorization patterns persist.

% \begin{figure*}[th]
%     \centering
%     % Left: Table in minipage
%     \begin{minipage}[t]{0.26\textwidth}
%         \vspace{0pt}

%     \end{minipage}%
%     \hfill
%     % Right: Figure in minipage
%     \begin{minipage}[t]{0.74\textwidth}
%         \vspace{0pt}
%         \centering
%         \includegraphics[width=\linewidth]{figures/post_patch_cross_experiment_comparison.pdf}
%         \caption{Comparison of models' 5-gram overlap ratio against the ground truth code snippet on the function reproduction task across benchmarks.}
%         \label{fig:func_reproduction_comparison}
%     \end{minipage}
% \end{figure*}

\begin{figure*}[th]
    \centering
        \includegraphics[width=\linewidth]{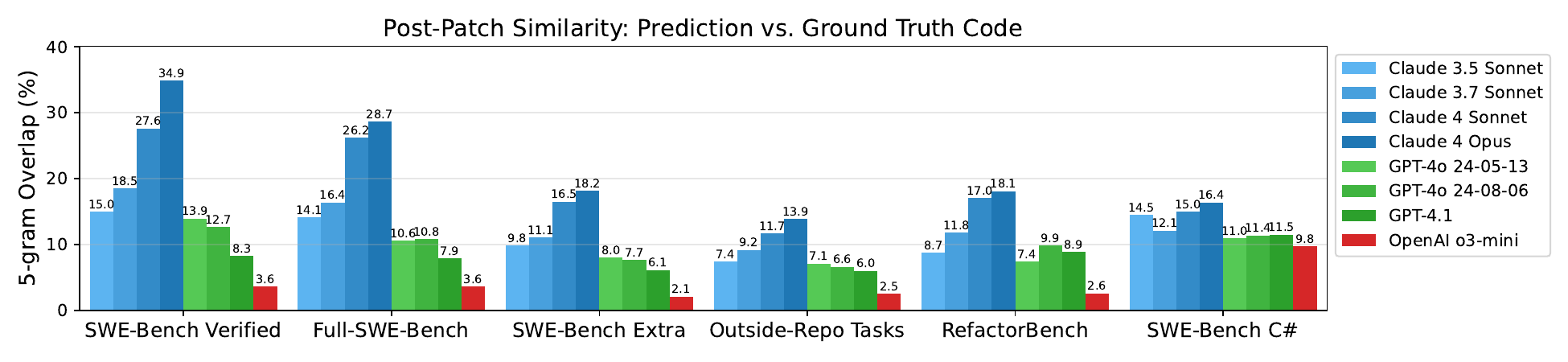}
        \caption{Comparison of models' 5-gram overlap ratio against the ground truth code snippet on the function reproduction task across benchmarks.}
        \label{fig:func_reproduction_comparison}
\end{figure*}

\begin{table}[H] % use H or !h to lock inside minipage
    \centering
    \footnotesize
    \begin{tabular}{l r}
    \toprule
    \textbf{Model} & \textbf{Match (\%)} \\
    \midrule
    Claude 3.5 Sonnet   & 12.1 \\
    Claude 3.7 Sonnet   & 12.3 \\
    Claude 4 Sonnet     & 21.4 \\
    Claude 4 Opus       & 31.6 \\
    GPT-4.1             & 17.4 \\
    GPT-4o 24-05-13     & 18.4 \\
    GPT-4o 24-08-16     & 18.2 \\
    OpenAI o3-mini      & 11.7 \\
    \bottomrule
    \end{tabular}
    \caption{Instance-level verbatim match percentage across models on SWE-Bench Verified.}
    \label{tab:verbatim_match}
\end{table}

% \begin{figure*}[th]
%     \centering
%     \includegraphics[width=0.72\linewidth]{figures/post_patch_cross_experiment_comparison.pdf}
%     \caption{Comparison of models' 5 gram overlap ratio comparing against the ground truth code snippet on the function reproduction task across benchmarks.}

%     \label{fig:func_reproduction_comparison}
% \end{figure*}
% \begin{table}[ht]
%     \centering
%     \scriptsize
%     \begin{tabular}{l r}
%     \toprule
%     \textbf{Model} & \textbf{Instance (\%)} \\
%     \midrule
%     Claude 3.5 Sonnet   & 12.1 \\
%     Claude 3.7 Sonnet   & 12.3 \\
%     Claude 4 Sonnet     & 21.4 \\
%     Claude 4 Opus       & 31.6 \\
%     GPT-4.1             & 17.4 \\
%     GPT-4o 24-05-13     & 18.4 \\
%     GPT-4o 24-08-16     & 18.2 \\
%     OpenAI o3-mini      & 11.7 \\
%     \bottomrule
%     \end{tabular}
%     \caption{Instance-level verbatim match percentage across models on SWE-Bench Verified.}
%     \label{tab:verbatim_match}
% \end{table}

\subsection{Function Reproduction Result}
We analyze model performance on the function reproduction task in this subsection.
As shown in Figure~\ref{fig:func_reproduction_comparison}, models achieve up to 34.9\% 5-gram overlap ratio reproducing ground truth functions from SWE-Bench Verified compared to up to 18.2\% on instances outside the SWE-Bench ecosystem\footnote{ Due to limited resources, we were only able to conduct experiments for \texttt{o3-mini} from among the OpenAI reasoning models.
}.
With limited context provided, only function names and issue descriptions, models lack sufficient information to systematically derive correct implementations. High consecutive 5-gram overlap under these constraints requires reproducing exact textual sequences, indicating reliance on memorized training data rather than reasoning.

\paragraph{Instance-Specific Memorization}: Performance reveals a clear hierarchy mirroring benchmark curation processes: SWE Verified (34.9\%) $>$ SWE Full (28.7\%) $>$ SWE Extra (18.2\%). Critically, SWE Extra performance drops to the same level as external benchmarks (RefactorBench: 18.1\%, Outside Tasks: 13.9\%), despite originating from identical repositories as Verified. This pattern indicates that models have specifically memorized the curated, canonical solutions that human evaluators deemed high-quality, rather than developing general familiarity with repositories.

These results demonstrate that SWE-Bench Verified performance substantially reflects memorization of training sequences. The inability to reproduce functions from less-curated instances within the same repositories, combined with performance equivalence between SWE Extra and external benchmarks, reveals that apparent programming ability might result from an artifact of exposure to the specific, high-quality solutions selected for benchmark inclusion.

\subsection{Direct Memorization on SWE-Bench Verified}

Table~\ref{tab:verbatim_match} presents the results of our prefix completion experiment across eight SoTA language models, showing substantial evidence of training data contamination in SWE-Bench Verified, with the instance-level verbatim match percentage ranging from 11.7\% to 31.6\% across evaluated models.

The Claude family exhibits a monotonic increase in memorization rates corresponding to model generation, rising from 12.1\% to 31.6\%. This progression suggests a strong correlation between model capacity and memorization. Notably, Claude 4 Opus demonstrates verbatim generation for nearly a third of all test instances.
In contrast, the GPT family displays relatively stable memorization rates (17.4\%-18.4\%), indicating potential differences in training data curation or architectural choices that affect memorization. OpenAI o3-mini exhibits the lowest repro rate at 11.7\%.

These results have significant implications for benchmark validity. The ability of models to reproduce exact code sequences when provided merely with contextual prefixes, without any problem description or bug identification, strongly suggests that performance on these instances reflects memorization rather than algorithmic reasoning or program comprehension. This finding strengthens our hypothesis that a substantial portion of reported performance gains on SWE-Bench may be attributed to memorization rather than genuine advances in programming capabilities.

\subsection{Comparison of 5-Gram Overlap for Buggy V.S. Ground Truth Code Snippet}
\begin{figure*}[th]
    \centering
    \includegraphics[width=0.9\linewidth]{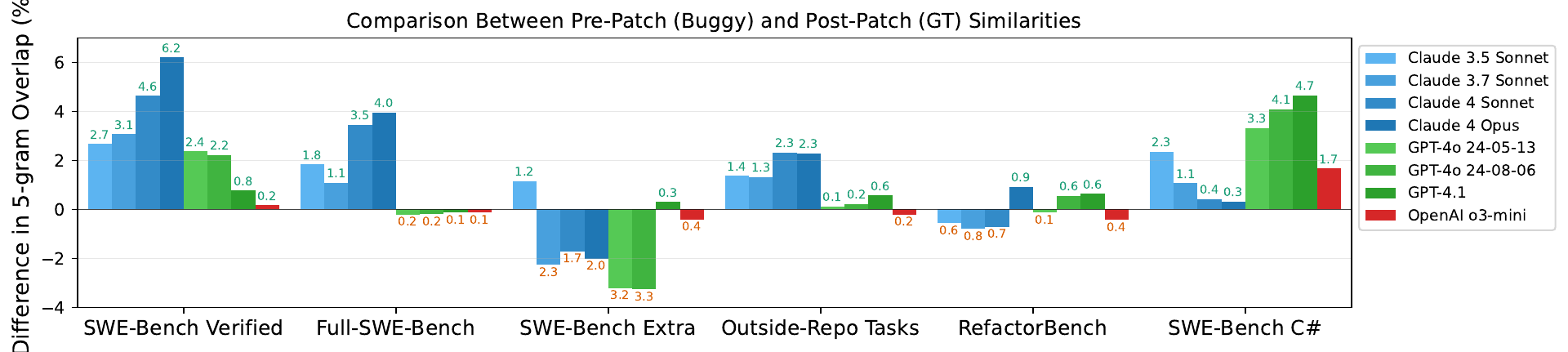}
    \caption{Difference in models’ 5-gram overlap ratio when comparing against the ground truth code snippet versus the buggy code snippet in the function reproduction task across benchmarks. A higher difference indicates that the model’s solution is more similar to the ground truth.}
    \label{fig:diff_5_gram_overlap}
\end{figure*}

We also report the difference of 5 gram overlap~\eqref{eq:delta5}, where $\text{overlap}_5(x,y)$ is the 5 gram overlap of $x$ and $y$. Positive values indicate the model output more closely matches the patched implementation; negative values indicate closer similarity to the buggy code.

\begin{equation}
\Delta_5 = \text{overlap}_5(\hat{f}, f_{\text{GT}}) - \text{overlap}_5(\hat{f}, f_{\text{buggy}})
\label{eq:delta5}
\end{equation}

Figure~\ref{fig:diff_5_gram_overlap} summarizes $\Delta_5$ across benchmarks and model families. Three broad patterns emerge.

\paragraph{(1) Provider‑level asymmetry on legacy benchmarks.} Anthropic's Claude models show consistently positive differentials on {SWE‑Bench Verified}, {Full-SWE-Bench}, and {Outside-Repo Tasks} (roughly +2 to -6 percentage points), suggesting that these models retain verbatim shards of the patched implementations. OpenAI models yield smaller positives (generally less than +2pp) over the same sets, indicating weaker memorization of those specific fixes.

\paragraph{(2) Freshness stress test.} All models, except Claude 3.5 Sonnet and GPT 4.1, produce \emph{negative} $\Delta_5$ on {SWE‑Bench Extra}. Because this split is mined largely from post‑cutoff commits, the patches are unlikely to be in training data. The systematic shift toward the buggy reference, therefore, supports our contamination‑mitigation claim: when the patched code is unseen, models do not memorize it from context.

\paragraph{(3) Language / ecosystem skew.} On the \textsc{SWE C\#} subset, OpenAI's GPT‑4* family exhibits the largest positive differentials (\textasciitilde{}+3 to 5pp), exceeding Claude. This may suggest heavier exposure to patched C\# / .NET code in OpenAI training data (or lighter deduplication of that ecosystem) compared to Anthropic.

Taken together, the function reproduction experiment provides a direct, contrastive probe of code memorization: when historical (patched) code is in a model's training mix, we observe measurable positive $\Delta_5$; when the code post‑dates the model (\textsc{SWE‑Bench Extra}), the signal disappears or reverses. Additionally, the high tendency towards the fixed version of code on SWE-Bench Verified and SWE-Bench C\# indicates a higher possibility of benchmark memorization on these two settings. These results underscore the necessity of freshness‑controlled benchmarks when interpreting coding model performance and memorization risk.

\section{Related Work}

\subsection{Coding Benchmarks and Agents} Various coding benchmarks have been proposed to evaluate LLMs and LLM agents', including SWE-Bench, BigCodeBench, SWE-Gym, EvoEval~\cite{jimenez2024swebench, related-bigcodebench, related-evoeval}. Among them, SWE-Bench has emerged as the most popular evaluation method for LLM's coding abilities \citep{jimenez2024swebench}. The benchmark consists of real GitHub issues and their corresponding patches, providing a seemingly realistic evaluation setting. However, the public nature of GitHub data raises concerns about potential exposure during training. Recent agent systems have shown dramatic improvements on SWE-Bench \citep{yang2024sweagent, wang2024opendevinopenplatformai}, but the extent to which these improvements reflect genuine problem-solving versus pattern matching remains unclear.

\subsection{Benchmark Contamination} The issue of benchmark contamination, where models are inadvertently trained on eval data, is a well-documented problem that can lead to inflated performance metrics without genuine understanding~\cite{zhou2023dontmakellmevaluation}. To combat this in the coding domain, prior work has developed two primary strategies: task mutation and metric-based probing.

\textbf{Task mutation approaches} aim to detect memorization by altering existing benchmarks. For example, EvoEval~\cite{related-evoeval} uses LLMs to create semantic-altering transformations of HumanEval~\cite{related-2021codex} problems, while other methods~\cite{related-memorizeorgeneralize} use AST-level mutations to measure how performance changes with structural code changes. Similarly, TaskEval~\cite{related-taskeval} measures prompt sensitivity by creating variations of a task. These methods are effective at revealing overfitting on specific problem formulations.

\textbf{Metric-based probing approaches}, in contrast, analyze model outputs without changing the task itself. These techniques~\cite{li2023estimatingcontaminationperplexityquantifying,related-llmmemorizebugbenchmarks} include using perplexity or other statistical measures like Negative Log Likelihood (NLL) and n-gram accuracy to infer if model has likely seen a specific bug or solution before. However, these measures cannot be applied to commercial models as we do not have access to the model's hidden states. N-gram similarity, while applicable, is a noisy indicator for complex code patches; any functionally correct solution will inevitably share text overlap with ground truth, making it hard to distinguish genuine problem-solving from verbatim memorization.

%While existing mutation and metric-based methods are effective for single-function code generation, they face limitations when applied to complex, multi-file software engineering benchmarks like SWE-Bench. Task mutation approaches require generating semantically equivalent variations of real-world GitHub issues, a significantly more challenging task than mutating isolated coding problems.

Our work introduces a novel cross benchmark analysis framework that addresses these limitations by comparing models across related benchmarks as memorization proxies. Rather than requiring knowledge of training data exposure, we systematically compare model performance across benchmark variants and external evaluations to detect suspicious performance patterns. This provides a systematic, reusable framework for detecting memorization in any coding benchmark without requiring access to training data or model internals.
\section{Conclusion}
This work presents evidence that current evaluations of LLM coding abilities may be compromised by benchmark contamination and memorization. Our systematic evaluation across ten models reveals two distinct memorization patterns: instance-specific, evidenced by graduated performance decay within SWE-Bench ecosystem (in both file-path identification and function reproduction tasks), and repository-bias memorization, demonstrated by substantial performance drops on external repos (up to 47 percentage points for file-path identification) despite their popularity. The consistency of these patterns across different model families and vendors indicates systematic exposure patterns of SWE-Bench Verified's data during training.

These findings suggest that the improvements in SWE-Bench Verified performance may partially reflect benchmark-specific optimization rather than genuine advances in coding capabilities. The field urgently needs more robust evaluation frameworks with temporal controls to prevent training data contamination, cross-repository validation to test generalization of these models beyond familiar codebases, and systematic cross-benchmark analysis to distinguish between memorization and transferable skills. Our work highlights the importance of developing contamination-resistant benchmarks and more sophisticated evaluation methodologies to ensure that reported progress reflects genuine advances in software engineering capabilities rather than overfitting on dataset-specific artifacts.

%%
%% The next two lines define the bibliography style to be used, and
%% the bibliography file.
\bibliographystyle{ACM-Reference-Format}
\bibliography{sample-base}

%%
%% If your work has an appendix, this is the place to put it.
\appendix

\end{document}